\documentclass[letterpaper, 10 pt, conference]{IEEEtran}
\IEEEoverridecommandlockouts
\usepackage{cite}
\usepackage{amsmath,amssymb,amsfonts}
\usepackage{algorithmic}
\usepackage{graphicx}
\usepackage{textcomp}
\usepackage{xcolor}
\usepackage{caption, subcaption}
\usepackage[ruled,vlined,linesnumbered]{algorithm2e}
\usepackage{array}

\usepackage{tikz}
\usetikzlibrary{calc}
\usepackage{pgfplots}
\usetikzlibrary{plotmarks}
\pgfplotsset{compat=newest}
\usetikzlibrary{patterns}
\usetikzlibrary{calc, pgfplots.dateplot}
\usetikzlibrary{external}
\usetikzlibrary{quotes, angles}

\usepackage{booktabs, multirow} %
\usepackage{soul}%

\newcolumntype{L}[1]{>{\raggedright\let\newline\\\arraybackslash\hspace{0pt}}m{#1}}
\newcolumntype{C}[1]{>{\centering\let\newline\\\arraybackslash\hspace{0pt}}m{#1}}
\newcolumntype{R}[1]{>{\raggedleft\let\newline\\\arraybackslash\hspace{0pt}}m{#1}}

\def\BibTeX{{\rm B\kern-.05em{\sc i\kern-.025em b}\kern-.08em
    T\kern-.1667em\lower.7ex\hbox{E}\kern-.125emX}}
\begin{document}

 \newcommand{\includetikz}[1]{
    \tikzsetnextfilename{./tikz/#1}
    \input{./tikz/#1.tikz}
}

\title{Vehicle Occurrence-based Parking Space Detection
\thanks{This work has been supported by the Brazilian National Council for Scientific and Technological Development (CNPq) -- Grant 405511/2022-1.}
}

\author{\IEEEauthorblockN{Paulo R. Lisboa de Almeida\IEEEauthorrefmark{1}, Jeovane Honório Alves\IEEEauthorrefmark{1}, Luiz S. Oliveira\IEEEauthorrefmark{1},\\ Andre Gustavo Hochuli\IEEEauthorrefmark{2}, João V. Fröhlich\IEEEauthorrefmark{3}, and Rodrigo A. Krauel\IEEEauthorrefmark{3}}
\IEEEauthorblockA{\IEEEauthorrefmark{1}Department of Informatics, Federal University of Paraná (UFPR), Curitiba, Brazil\\
Email: paulorla@ufpr.br, jeohalves@gmail.com, luiz.oliveira@ufpr.br}
\IEEEauthorblockA{\IEEEauthorrefmark{2}Graduate Program in Informatics, Pontifical Catholic University of Paraná (PUCPR), Curitiba, Brazil\\
Email: aghochuli@ppgia.pucpr.br}
\IEEEauthorblockA{\IEEEauthorrefmark{3}Department of Computer Science, Santa Catarina State University (UDESC), Joinville, Brazil\\
Email: joaovitorfrohlich@gmail.com, rodrigoakrauel@gmail.com}}

\maketitle

\begin{abstract}
Smart-parking solutions use sensors, cameras, and data analysis to improve parking efficiency and reduce traffic congestion. Computer vision-based methods have been used extensively in recent years to tackle the problem of parking lot management, but most of the works assume that the parking spots are manually labeled, impacting the cost and feasibility of deployment. To fill this gap, this work presents an automatic parking space detection method, which receives a sequence of images of a parking lot and returns a list of coordinates identifying the detected parking spaces. The proposed method employs instance segmentation to identify cars and, using vehicle occurrence, generate a heat map of parking spaces. The results using twelve different subsets from the PKLot and CNRPark-EXT parking lot datasets show that the method achieved an AP25 score up to 95.60\% and AP50 score up to 79.90\%.
\end{abstract}

\begin{IEEEkeywords}
Smart-parking, Parking Space Detection, Deep Learning, Instance Segmentation. 
\end{IEEEkeywords}

\section{Introduction}

Smart-parking solutions use sensors, cameras, and data analytics to optimize parking spaces. The goal is to reduce traffic congestion and make finding a parking spot more efficient. Displaying real-time information about the availability of parking spots to drivers, either through a smartphone app or other means, is a key feature of smart parking solutions.

In the last decade, extensive research has been conducted utilizing computer vision-based methods to tackle the problem of parking lot management. In this context, public datasets have been proposed \cite{almeidaEtAl2015, amatoEtAl2016, nietoEtAl2019} and several different approaches ranging from shallow to deep learning have been reported to detect and classify parking spaces. Systematic reviews can be found in \cite{smartcities4020032} and \cite{Almeida2022}.

As pointed out in \cite{Almeida2022}, most of the works published in the literature assume that the parking spots are manually labeled, and all the system needs to do is to indicate if there is a car in that parking spot. However, the difficulty and time required for labeling images can impact the cost and feasibility of deploying computer vision systems.

\begin{figure}[ht!]
\centering
\includegraphics[width=4.6cm]{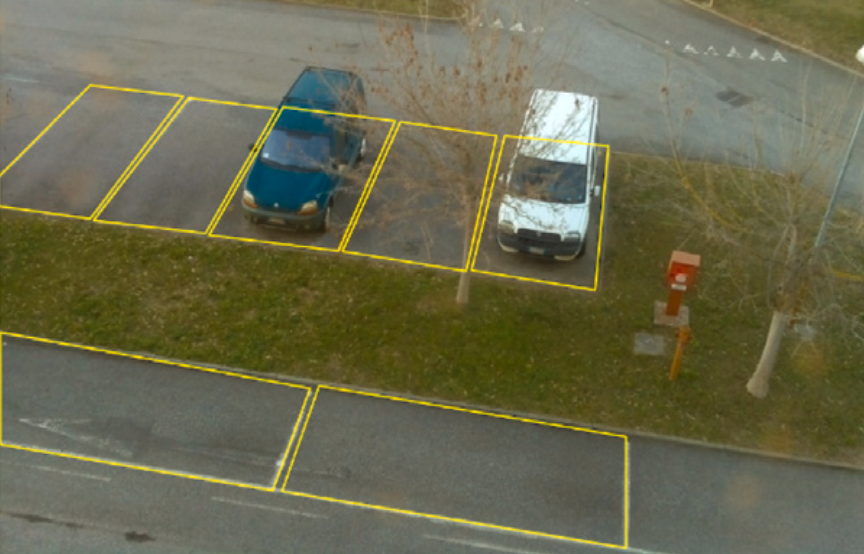}
\caption{Parking space locations (image from CNRPark-EXT).}
\label{fig:datasets}
\end{figure}

In this scenario, automatic parking space detection is mandatory to facilitate the deployment of smart parking solutions. 
This task is depicted in Figure \ref{fig:datasets}, where the yellow coordinates of the rotated rectangles (defining each spot) should be automatically defined. 
It is a challenging task since parking spaces are similar to roads, i.e., how can a model discriminate between a parking space and a road segment? The presence of cars may hinder correct detection, especially for methods that rely on the painted demarcations in the parking lots that delimit the parking spaces.

\vspace{-0.3cm}

The literature shows some efforts towards automatic parking space detection, which can be classified into classical image processing methods \cite{bohushEtAl2018,ZhangEtAl2019}, grid-based methods \cite{nietoEtAl2019,vitekMelnicuk2018}, and deep learning \cite{Li2017,padmasiriEtAl2020,Kirtibhai2020}. As pointed out in \cite{Almeida2022}, most works did not address the problem directly but discussed it as an intermediary step for the classification system. Consequently, they do not provide quantitative results, making it impossible to compare different works. 

The main contribution of this work is a method specific to automatic parking space detection, which receives as input a sequence of images of a parking lot and returns a list of coordinates of rotated rectangles identifying the detected parking spaces.  The rationale is based on the premise that parking spaces are regions where vehicles remain stationary for extended periods. The proposed method employs instance segmentation to identify cars, which is then utilized to generate an automatic heat map of parking spaces.

Through comprehensive experimental results on twelve parking lot subsets from the PKLot and CNRPark-EXT datasets, we demonstrate that the proposed method can detect parking spaces without prior knowledge, avoiding the laborious task of manual segmentation. The results show that the method achieved an average precision at an Intersection over Union (IoU) threshold of 25\% (AP25) up to 95.60\%, and an average precision at an IoU threshold of 50\% (AP50) up to 79.90\%. As discussed in our experiments, these metrics can be impacted by illegal parking and unused parking spaces. Although this affects the AP metric, it highlights the method's adaptability to dynamic parking scenarios, such as seasonal events, without prior knowledge of demarcations. Besides, our results compare favorably to the literature.
\section{Related Works}

Bohush et al. \cite{bohushEtAl2018} utilized classical image processing techniques for automatic parking extraction by applying a perspective transformation to make parking spots rectangular and parallel to the axes. They detected the painted lines in parking lots using Otsu's binarization and morphological operations. Similarly, Zhang et al. \cite{ZhangEtAl2019} applied a perspective transformation and used classical image processing techniques such as Canny and Gaussian edge detectors for parking space detection. No quantitative results were made available in both \cite{bohushEtAl2018} and \cite{ZhangEtAl2019}.

In \cite{ZhangEtAl2020}, classical image processing methods were used for automatic parking space extraction. The Canny edge detector and Hough transform were used to detect the lines defining each parking spot. However, only images of an empty parking lot with visible lines were used, and a private dataset was employed. No quantitative results were presented.

In \cite{vitekMelnicuk2018}, a grid-based approach was used where Histogram of Oriented Gradients' (HOG) features were extracted from grid blocks and classified as either cars or streets using a classifier. In the merging phase, blocks classified as cars were merged into single parking spaces. However, it's important to note that the approach only detects cars, not parking spaces, as a car may be just passing by the parking lot. The authors only provided qualitative results. Similarly, in \cite{nietoEtAl2019}, the authors assumed that the car park area was rectangular and formed a parking grid. Using a homography matrix and a regular image captured by a camera, they calculated the equivalent aerial image of the parking lot. Based on the grid's corner coordinates and the number of rows and columns, they could automatically define the parking spaces, but no quantitative results were provided.

In \cite{agrawalEtAl2020} and \cite{ColeiroEtAl2020}, the Mask R-CNN network \cite{kaimingEtAl2017} was used to identify cars. In \cite{agrawalEtAl2020}, the authors used the identified cars to extract parking spaces by assuming areas where cars stay parked for long periods are parking spaces. Similarly, \cite{ColeiroEtAl2020} used the Mask R-CNN to extract car bounding boxes and generated a heat map to detect stationary car areas, only providing qualitative results in a private dataset for parking spot segmentation. In this vein, the work proposed in \cite{Karthick2021} presents a technique that aims to enhance car detection by improving the lighting conditions and incorporating models that can detect vehicles from different angles. However, the aforementioned works lack in-depth discussions and quantitative results regarding the detection of parking spaces.

In \cite{padmasiriEtAl2020}, the ResNet \cite{heEtAl2016}, and Faster-RCNN \cite{renEtAl2015} networks were used to detect parking spaces, reporting results using the AP50 metric and PKLot dataset. However, it's important to note that the authors seem to have used images from the same parking lot for both training and testing, which may lead to biased results.

In \cite{hurst2020robust}, the authors use satellite imagery to perform automated extraction of parking blocks. A parking block is defined as a contiguous set of parking spaces, discernible in satellite images taken with an orthogonal orientation to the parking lot and featuring visible lines that delimit individual parking spaces. Camera parameters of the parking lot are required to match the satellite images with parking lot surveillance camera images. The authors use a U-Net architecture for parking space detection. Additionally, the authors introduce the APKLOT dataset, comprising 500 satellite images of parking lots. An Intersection over Union (IoU) score of 62.1 on the APKLOT dataset was reported.

The authors in \cite{Kirtibhai2020} and \cite{patelMeduri2020} present an approach where the Faster R-CNN or YOLOv4 is used for car detection. The car's bounding boxes detected in two or three consecutive frames are compared to detect whether a vehicle is stationary. The stationary car's positions (a non-rotated bounding box) is deemed as a parking space. The authors evaluated the proposed approach in three busy days from CNRPark-EXT for each weather condition available.

Table \ref{table:comparacaoMetodosContagem} summarizes the works analyzed in this section, emphasizing the need for improved automatic parking space detection methods and the importance of presenting quantitative results. For performance assessment, well-established metrics such as F1-score, IoU, or AP \cite{everinghamEtAl2010} can be used. The chosen metrics should not depend on true negatives as the entire parking lot, excluding the parking space polygons, is considered a true negative. Metrics that rely on true negatives, like accuracy, may result in biased results.

\begin{table}[htpb]
\caption{Automatic parking space detection approaches overview.}
\centering 
    \setlength{\tabcolsep}{3pt}
    \begin{tabular}[c]{L{0.6cm}L{2.5cm}L{1.0cm}L{1.9cm}L{1.8cm}}
        \hline
        Ref & Approach& Painted lines?& Results (Metric) & Datasets Used \\
        \hline

        \cite{Kirtibhai2020} & Faster-RCNN & no & 83.1 (AP50) & CNRPark-EXT\\

        \cite{patelMeduri2020} & YOLOv4 & no & 97.6 (AP50) & CNRPark-EXT\\
        
        \cite{bohushEtAl2018} & Image processing & yes & - & PKLot\\
        
        \cite{vitekMelnicuk2018} & Grid blocks classification & no & - & PKLot, Private Data\\
        \cite{nietoEtAl2019} & Homography matrices & no & - & PLds\\
        
        \cite{ZhangEtAl2019} & Edge detectors & yes & - & PKLot\\

        \cite{agrawalEtAl2020} & Mask R-CNN  & no & - & PKLot\\
        
        \cite{hurst2020robust} & Satellite images and U-Net & yes & 62.1 (IoU) & APKLOT\\
        
        \cite{padmasiriEtAl2020} & ResNet and Faster-RCNN & no & 63.6 (AP50) & PKLot\\
        
        \cite{ZhangEtAl2020} & Canny Edge Detector & yes & - & Private Data\\
        
        \hline
        
    \end{tabular}
\label{table:comparacaoMetodosContagem}
\end{table}

\section{Proposed Method}\label{sec:proposed_method}

\begin{figure}[hbt!]
\centering
\includegraphics[width=\linewidth]{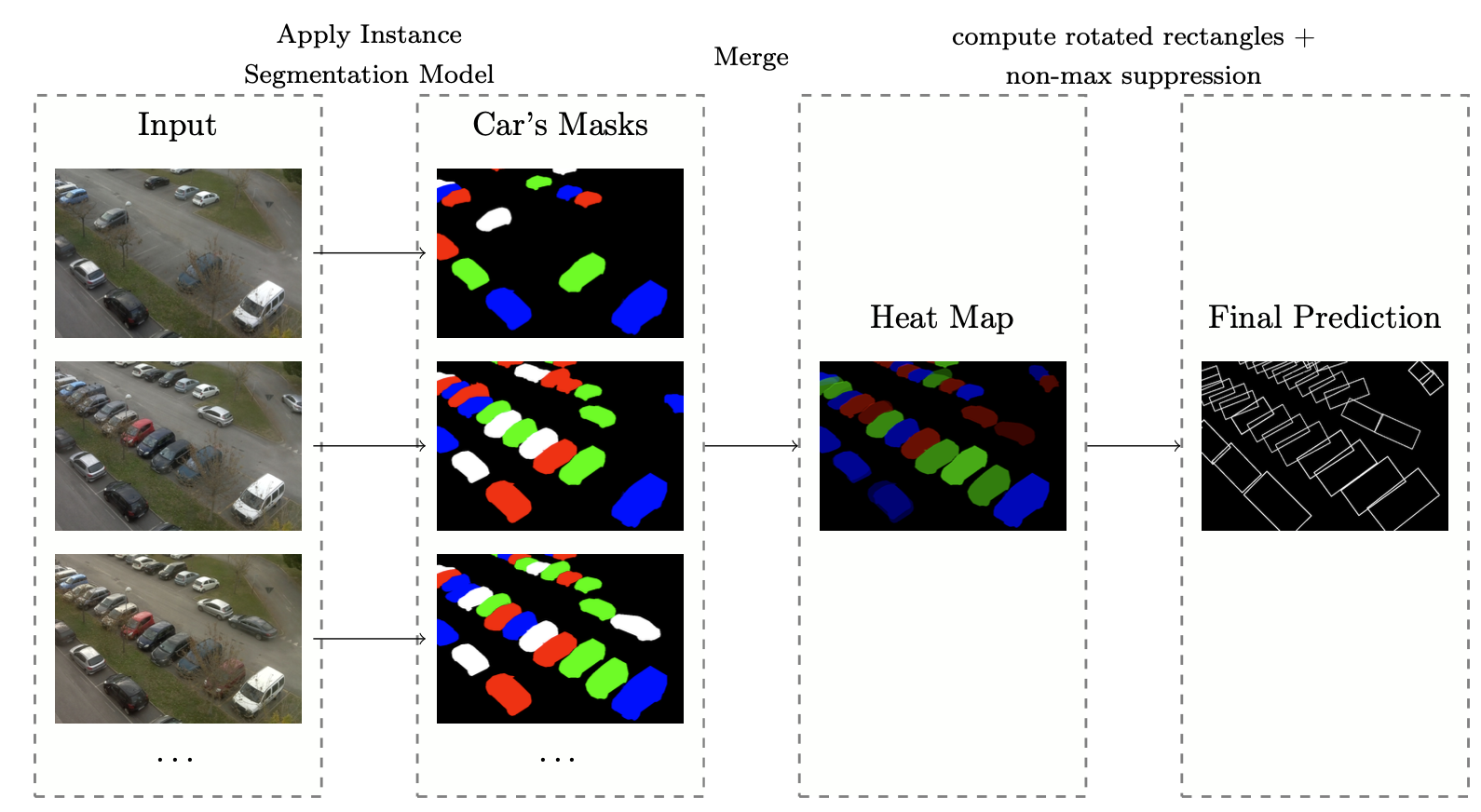}
\caption{Proposed method}
\label{fig:method}
\end{figure}

The proposed method is shown in Figure \ref{fig:method}. It operates under the premise that parking spaces are regions where vehicles remain stationary for extended periods. The method creates heat maps that depict areas in the images where cars are frequently detected. This approach is similar to \cite{ColeiroEtAl2020}. However, it is important to note that the authors in \cite{ColeiroEtAl2020} employ non-rotated bounding boxes for parking space representation and do not provide details on how the heat map is processed to segment the parking spaces.

To generate a heat map of the utilization of a parking lot, images or a video of a full day of operation are required. This duration of data acquisition is considered sufficient for the analysis and can be readily obtained through the system's deployment. Our approach requires only the images, without additional annotations or preprocessing, since it focuses on identifying areas where vehicles tend to park. It is recommended that images be taken at regular intervals, with a suggested interval of 5 minutes, to accurately capture changes in vehicle movements and turnover within the parking lot. Although, good results have been achieved with a 30-minute interval in the CNRPark-EXT dataset.

\vspace{-0.3cm}

Our study employed an instance segmentation model (ISM) to segment individual cars in each image. Specifically, we utilized the Cascade Mask R-CNN \cite{kaimingEtAl2017} (extension of Cascade R-CNN \cite{cai2018cascade}), adding a mask head to the model, similar to the Mask R-CNN approach. As for the backbone of the Cascade Mask R-CNN, we used the ConvNeXt-S model \cite{liu2022convnet}, since, combined with the Cascade Mask R-CNN, it has state-of-the-art results in popular benchmarks for instance segmentation. The model was pre-trained on the COCO dataset.

The segmented car images from a day of parking lot operation are integrated to create a heat map. For each region of the heat map, an a posteriori probability was computed, representing the probability of the region being a parking space.

\begin{algorithm}[htpb]
	\footnotesize
	\SetKwInput{KwData}{Input}
	\KwData{Images from one operation day ($IM = \{I_1,I_2,\dots,I_n\}$),\\
	    Trained instance segmentation model $(ISM)$,\\
	    Minimum IoU to sum the images $t_{sum}$\\
	    Non-maximum suppression threshold $t_{nms}$\\
	}
	\KwResult{Set containing the Predictions $(R)$}
	
	$P \leftarrow \varnothing \hspace{0.2cm} $\tcp{P holds the masks}
	$R \leftarrow \varnothing \hspace{0.2cm} $\tcp{R holds the resulting spaces}
	
	\ForEach{Image $i \in IM$}{
	    $L \leftarrow getCarsMasks(ISM, i)$\\
	    \tcp{Merge consecutive masks to create a heat map}
	    \ForEach{Mask $m \in L$}{
	        \tcp{Find the mask in P with the biggest IoU with m}
	        $k \leftarrow argMaxIoU(P,m)$\\
	        \If{$boundingBoxIoU(m,k) \geq t_{sum} $}{
	            \tcp{Car parked in an existing position}
	            $k \leftarrow k + m$ \hspace{0.2cm} \tcp{sum the images}
    		}\Else{
    		    \tcp{Add the mask as a new image in the set P}
    		    $P \leftarrow P \cup Mask$
    		}
	    }
    }
    \tcp{Rotated rectangles and a posteriori computation}
    \ForEach{Image $j \in P$}{
            $r.rotatedRect \leftarrow rotatedBoundingRectangle(j)$\\
            $r.aposteriori \leftarrow \frac{sumPixelVals(j)/numNomZeroPixels(j)}{numImages(IM)}$\\
            $R \leftarrow R \cup r$\\
    }
    $R \leftarrow nonMaxSupression(R, t_{nms})$\\
    $\Return\ R$
\caption{Parking space detection algorithm.}
\label{alg:ParkingDetection}
\end{algorithm}

Algorithm \ref{alg:ParkingDetection} outlines the complete parking space detection procedure, where we assume that each car mask identified by the ISM is represented as a matrix with the same dimensions as the original parking lot image. This matrix contains zeros in areas not occupied by a car and ones in areas occupied by a car. If the ISM detects $n$ cars in an image, there must be $n$ such matrices computed and stored in the list $L$. The matrices are summed if a car is detected in the same position as previous cars. By summing the images of identified cars, we are creating a heat map. The reasoning behind this approach is that the parked vehicles will produce hotter zones than regions with moving traffic.

In the loop of line 11, we compute the a posteriori probability of each identified region being a parking space. This computation is based on the proportion of pixels with a car parked in each image. This value can be later used to accept/reject parking spaces based on the a posteriori generated for it (in a similar fashion as we do when using the a posteriori probability of a classifier to classify some instance).
\section{Experiments and results}

\subsection{Datasets}

\begin{figure}
    \centering
    \begin{subfigure}[b]{0.47\columnwidth}
        \centering
        \includegraphics[width=\columnwidth]{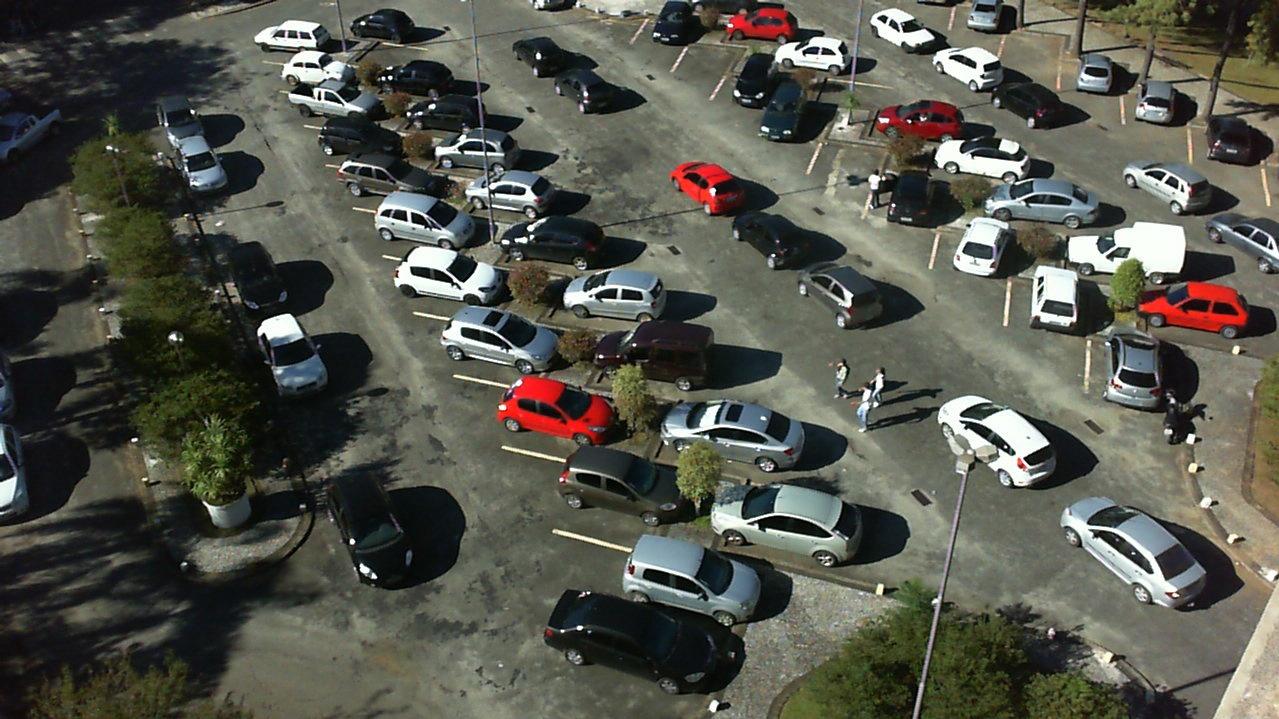}
        \caption{}                        
    \end{subfigure}    
    \hskip\baselineskip
    \begin{subfigure}[b]{0.35\columnwidth}  
        \centering 
        \includegraphics[width=\columnwidth]{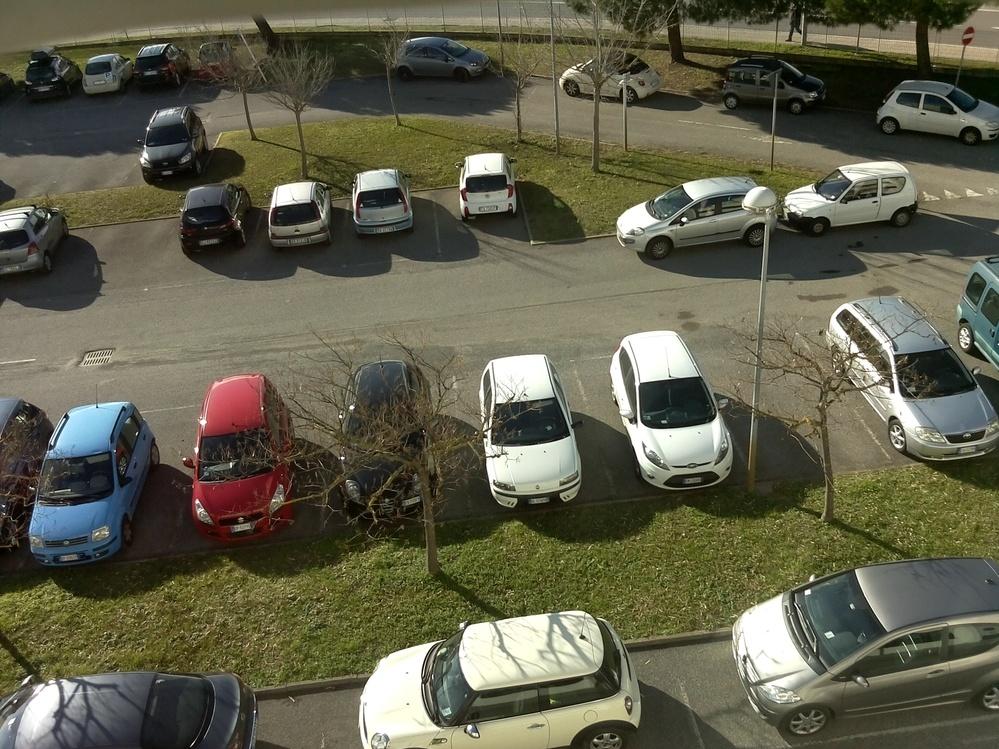}
        \caption{}                        
        \label{fig:mean and std of net24}
    \end{subfigure}
    \caption{Examples of cameras used in our experiments: (a) UFPR05 from PKLot and (b) Camera3 from CNRPark-EXT}
    \label{fig:cameras}
\end{figure}

To validate the proposed method, we used the PKLot\cite{almeidaEtAl2015} and CNRPark-EXT \cite{amatoEtAl2017} datasets, which are publicly available. These include UFPR04, UFPR05 (Figure \ref{fig:cameras}a), and PUCPR from the PKLot dataset, which comprises 3791, 4152 and, 4117 images, respectively, taken from 43, 46, and 134 parking spots, during varying weather conditions. Additionally, we used the Cameras 1 to 9 (see Figure \ref{fig:cameras}b for an example) from the CNRPark-EXT dataset, which consists of 458, 462, 452, 451, 461, 406, 406, 459, and 471 images, respectively from 37, 10, 23, 39, 48, 48, 46, 55, and 29 spots. We manually annotated the cars in both datasets (and some extra parking spaces for the PKLot dataset). As for the CNRPark-EXT, as the parking spaces were originally squared boxes containing a part of the parking spaces, new annotations polygons were created to include the entire parking space region. For the parking space detection stage, we removed non-working days (e.g., days when the parking lots were closed).

\subsection{Experiment Settings}

First, for the optimizer, we adopted AdamW \cite{loshchilov2017decoupled} with a learning rate of 3e-5 combined with the cosine annealing scheduler \cite{loshchilov2016sgdr}. The data augmentation used in the training step is horizontal and vertical random flip ($prob=0.5$), CLAHE (Contrast Limited Adaptive Histogram Equalization \cite{pizer1987adaptive}) using a grid size of 8 $\times$ 8, random size ([0.8,1.2] of the original image size), and random brightness, contrast, saturation and hue (parameter of $0.2$ for each transformation). For testing, only CLAHE is applied.

Online training (i.e., batch size $=1$) is employed. The Cascade Mask-RCNN was fine-tuned for 54 epochs when using the subsets of CNRPark-EXT since they have a similar number of images. For a similar number of iterations, UFPR04, UFPR05, and PUCPR were fine-tuned for 6, 7, and 6 epochs, respectively. 
We trained models only with images of one subset to assess the capability of detecting parking spaces of non-seen parking lots with minimal effort. Thus, twelve models were trained separately, three for PKLot and nine for CNRPark-EXT, each for a subset described previously. In Algorithm \ref{alg:ParkingDetection}, we used the parameter $t_{sum} = 0.5$. The parameter $t_{nms}$, used to compute the non-maximum suppression of the parking spaces, was set to $0.4$.

The experiments are described in two stages: car instance segmentation and parking space detection. For better understanding, experiments were executed with models trained and tested in different evaluations according to \cite{Almeida2022}: (1) ANGLE CHANGE -- same parking lot but with different angles (between subsets from the CNRPark-EXT dataset and between UFPR04 and UFPR05), (2) PARKING LOT CHANGE -- same dataset but different parking lots (UFPR04 and UFPR05 against PUCPR, and vice-versa), and (3) DATASET CHANGE -- between different datasets (PKLot subsets against the CNRPark-EXT subsets, and vice-versa).

For evaluation, we use the widely known Average Precision (AP) metric \cite{everinghamEtAl2010} at an IoU of 50\% (AP50) for the car instance segmentation step and IoU of 25\% (AP25) and 50\%  (AP50) for the parking space detection. Table \ref{tab:mean_ap} reports the average between models for each testing subset. For example, the UFPR04 subset's result in the DATASET CHANGE evaluation is the average between each of the nine models trained on the subsets of the CNRPark-EXT dataset. Finally, for each evaluation, we averaged the results of each testing subset for PKLot and CNRPark-EXT to obtain an overall result for both datasets (except in the PARKING LOT CHANGE evaluation, only the PKLot result is calculated since there is only one parking lot in the CNRPark-EXT).

\begin{table}[htpb]
\centering
\caption{Mean AP and standard deviation for the three proposed evaluations. Best subset's result for each evaluation are in bold. Overall dataset results are also depicted.
}
\label{tab:mean_ap}
\renewcommand{\arraystretch}{0.95}
\begin{tabular}{@{}lrrp{0.3cm}rrrr@{}}
\toprule
Testing subset & \multicolumn{2}{c}{Car} & \multicolumn{1}{c}{} & \multicolumn{4}{c}{Parking Space} \\ 
 & \multicolumn{2}{c}{Segmentation} & \multicolumn{1}{c}{} & \multicolumn{4}{c}{Detection} \\ \midrule
 & \multicolumn{2}{c}{AP50} &  & \multicolumn{2}{c}{AP25} & \multicolumn{2}{c}{AP50} \\ \midrule
\multicolumn{1}{c}{} & \multicolumn{1}{c}{mean} & \multicolumn{1}{c}{std} &  & \multicolumn{1}{c}{mean} & \multicolumn{1}{c}{std} & \multicolumn{1}{c}{mean} & \multicolumn{1}{c}{std} \\ \midrule
\multicolumn{8}{c}{ANGLE CHANGE} \\ \midrule
UFPR04 & \multicolumn{2}{c}{84.90} &  & 66.19 & 6.24 & 58.99 & 6.49 \\
UFPR05 & \multicolumn{2}{c}{\textbf{98.90}} &  & 85.32 & 5.55 & 53.40 & 10.83 \\
Camera1 & 82.69 & 1.85 &  & 79.42 & 4.61 & 28.40 & 6.84 \\
Camera2 & 94.94 & 2.45 &  & 84.10 & 7.75 & 70.15 & 11.11 \\
Camera3 & 92.28 & 3.53 &  & 88.54 & 4.48 & 71.98 & 9.16 \\
Camera4 & 96.68 & 1.92 &  & 94.00 & 3.72 & 57.51 & 11.20 \\
Camera5 & 96.80 & 1.31 &  & 92.42 & 2.94 & 40.28 & 5.23 \\
Camera6 & 96.36 & 1.83 &  & 90.51 & 3.89 & 45.60 & 6.42 \\
Camera7 & 95.24 & 2.70 &  & 92.46 & 4.09 & \textbf{79.90} & \textbf{8.24} \\
Camera8 & 96.58 & 1.98 &  & \textbf{95.60} & \textbf{4.12} & 64.19 & 7.04 \\
Camera9 & 92.60 & 2.96 &  & 83.95 & 5.66 & 53.72 & 11.00 \\ \midrule
\textbf{Overall results} & & & & & & & \\ \midrule
PKLot & 91.90 & 9.90 &  & 75.76 & 13.53 & 56.20 & 3.95 \\
CNRPark-EXT & 93.79 & 4.50 &  & 89.00 & 5.43 & 56.86 & 16.63 \\ \midrule
\multicolumn{8}{c}{PARKING LOT CHANGE (same dataset, different parking lots)} \\ \midrule
PUCPR & 81.65 & 0.64 &  & \textbf{90.06} & \textbf{4.61} & \textbf{78.48} & \textbf{4.64} \\
UFPR04 & \multicolumn{2}{c}{81.90} &  & 64.06 & 7.70 & 54.47 & 6.82 \\
UFPR05 & \multicolumn{2}{c}{\textbf{93.90}} &  & 77.22 & 4.61 & 44.39 & 9.10 \\ \midrule
\textbf{Overall results} & & & & & & & \\ \midrule
PKLot & 85.82 & 7.00 &  & 77.12 & 13.00 & 59.11 & 17.51 \\\midrule
\multicolumn{8}{c}{DATASET CHANGE} \\ \midrule
PUCPR & 76.07 & 6.11 &  & 86.00 & 7.50 & 64.73 & 9.23 \\
UFPR04 & 79.99 & 1.73 &  & 61.90 & 6.89 & 47.72 & 7.46 \\
UFPR05 & \textbf{94.73} & \textbf{0.95} &  & 81.72 & 5.24 & 48.42 & 9.32 \\
Camera1 & 77.30 & 7.39 &  & 79.88 & 5.47 & 28.29 & 7.63 \\
Camera2 & 79.90 & 16.84 &  & 75.36 & 13.99 & 54.07 & 20.32 \\
Camera3 & 81.83 & 5.45 &  & 81.97 & 7.09 & 57.67 & 9.94 \\
Camera4 & 91.97 & 2.57 &  & 89.71 & 5.24 & 44.05 & 9.37 \\
Camera5 & 91.57 & 2.78 &  & 89.49 & 3.64 & 32.40 & 5.84 \\
Camera6 & 89.17 & 3.21 &  & 86.90 & 4.96 & 32.83 & 7.17 \\
Camera7 & 89.83 & 2.97 &  & 92.02 & 4.95 & \textbf{69.10} & \textbf{9.50} \\
Camera8 & 92.10 & 2.21 &  & \textbf{92.62} & \textbf{4.67} & 52.89 & 7.95 \\
Camera9 & 88.30 & 2.97 &  & 84.08 & 6.48 & 48.29 & 10.77 \\ \midrule
\textbf{Overall results} & & & & & & & \\ \midrule
PKLot & 83.60 & 9.84 &  & 76.54 & 12.86 & 53.63 & 9.62 \\
CNRPark-EXT & 86.89 & 5.67 &  & 85.78 & 5.88 & 46.62 & 13.51 \\ \bottomrule
\end{tabular}
\end{table}

\subsection{Car Instance Segmentation}\label{sec:exp_car_instance}

In this step, we evaluate the results generated by the Cascade Mask-RCNN model for instance segmentation of cars. The resulting car instance masks are further used for the parking space detection step (Section \ref{subsec:parking_space_detection}).
In the \textit{Car Segmentation} column of Table \ref{tab:mean_ap}, we report the AP50 for every testing subset for the three different evaluations. Average and standard deviation are reported when multiple models' outputs are evaluated. When only one trained model was used to evaluate a testing subset, a single AP50 value is reported.

As expected, training with images from the same parking lot but with different camera angles shows superior results compared to the other two evaluation types. Although from the same parking lot, UFPR04 had inferior results than UFPR05 which, by a qualitative analysis, was mainly caused by object occlusion (e.g., trees occluding parked cars). Camera2 had the highest standard deviation, mainly because of the poor results when using the PUCPR model (AP50 of 60.50\%). We hypothesize that the different features between the two subsets are the cause, since PUCPR as the testing subset with the Camera2 trained model also had poor results compared to the remaining models (AP50 of 60.30\%).

Camera1 had the lowest results from CNRPark-EXT. We hypothesize that this was caused mainly by the upper-left part of Camera1, where there are more occluded and faraway parked cars, where the model would segment two cars as one, for example. In this case, treating the occlusion problem and employing strategies to correctly segment faraway adjacent cars would improve overall AP for these subsets.

\subsection{Parking Space Detection}
\label{subsec:parking_space_detection}

As discussed in Section \ref{sec:proposed_method}, this experiment assesses the proposed method for automatically detecting parking spaces. In this step, the Cascade Mask-RCNN's are used as the \textit{ISM} parameter in the Algorithm \ref{alg:ParkingDetection}, generating the segmented car's instances for the parking spaces detection step (see line 4 in Algorithm \ref{alg:ParkingDetection}). More specifically, for each day, the entire day's images are employed for assessing the parking spaces. Note that no test images were used in the training step.
The results reported in the column \textit{Parking Space Detection} of Table \ref{tab:mean_ap} allow us to draw some conclusions. Average and standard deviation of the AP metric are calculated by day.

First, Camera1 had the lowest AP50 score among all subsets. As stated in the previous section, this subset has strong angulation, with cars in the upper-left part of the image being far away. The method had difficulties to correctly segmenting the cars and also struggled to generate the rotated rectangle (i.e., parking space) in the correct orientation (i.e., angle). A similar problem is found in Cameras 4, 5, 6, and 8 of the CNRPark-EXT dataset. Differently from Camera1, these subsets are more vertically inclined (but the cameras were closer to the ground, unlike PKLot subsets). However, our approach had good results with the AP25 in the CNRPark-EXT dataset, with an overall score of 89\% and 85.78\% in the ANGLE CHANGE and DATASET CHANGE evaluations, showing that, though with lower IoU, our approach can effectively detect the parking spaces.

\begin{figure}[htbp]
    \centering
    \begin{subfigure}[b]{0.14\textwidth}
        \centering
        \includegraphics[width=\textwidth]{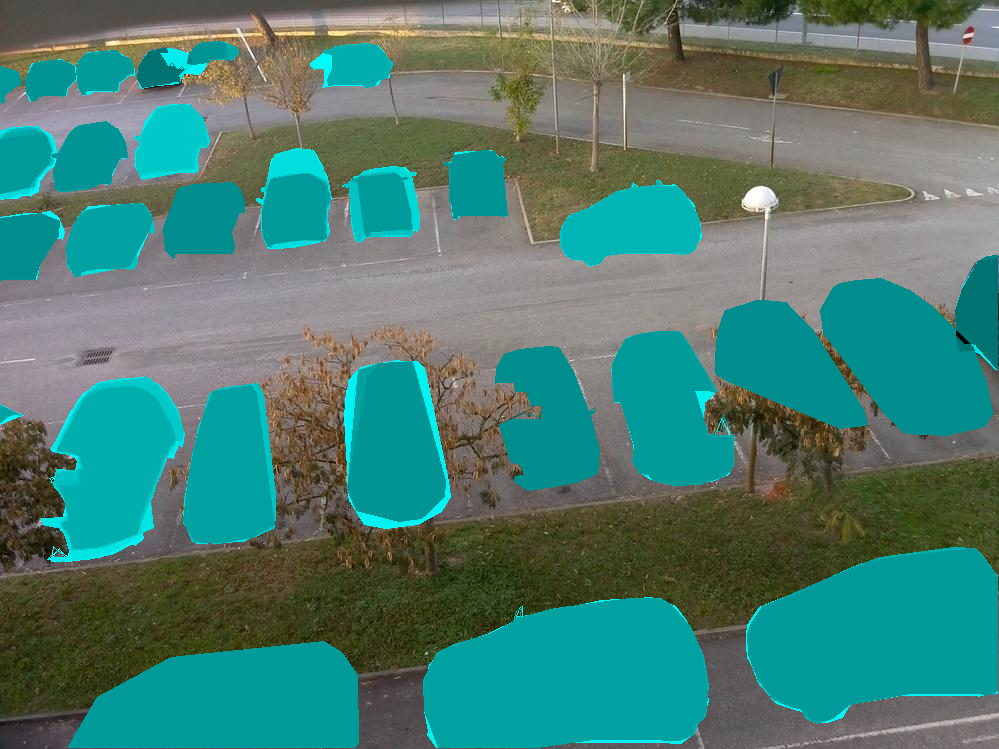}
        \caption{}                        
    \end{subfigure}            
    \begin{subfigure}[b]{0.14\textwidth}  
        \centering 
        \includegraphics[width=\textwidth]{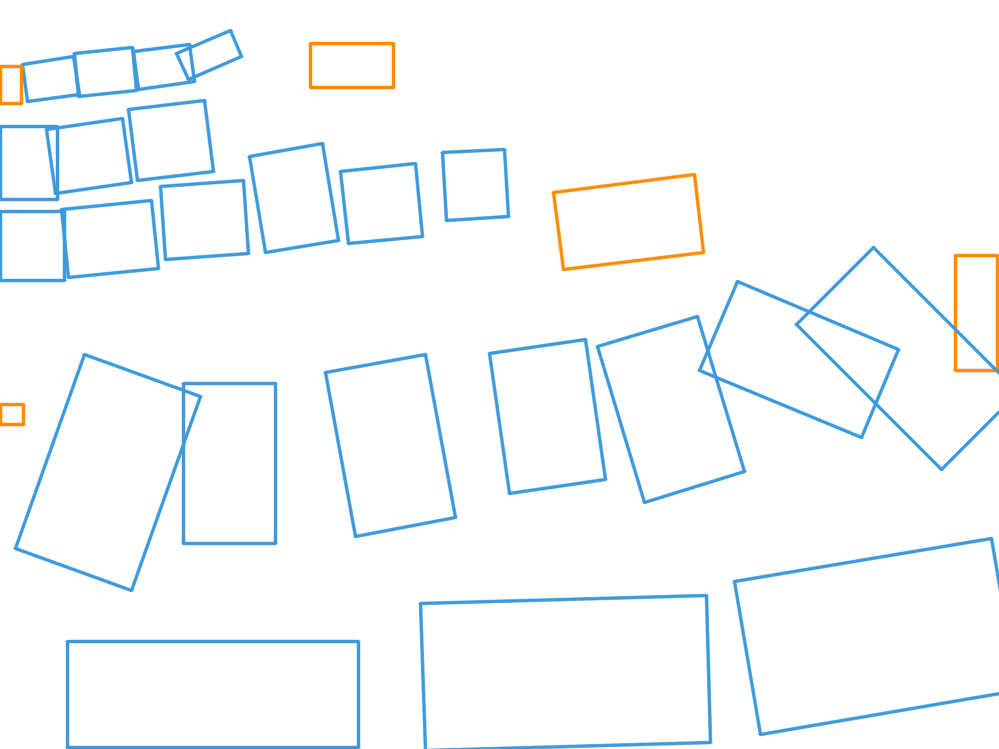}
        \caption{}                              
    \end{subfigure}    
    \begin{subfigure}[b]{0.14\textwidth}   
        \centering 
        \includegraphics[width=\textwidth]{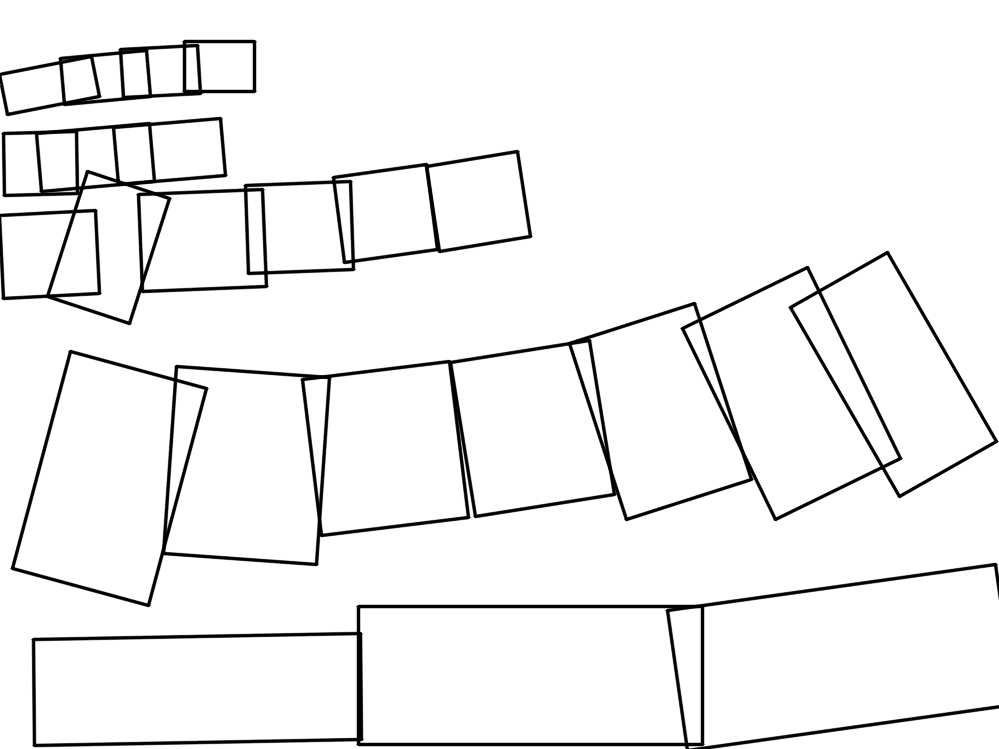}
        \caption{}
    \end{subfigure}    

    \vskip\baselineskip
    \begin{subfigure}[b]{0.14\textwidth}
        \centering
        \includegraphics[width=\textwidth]{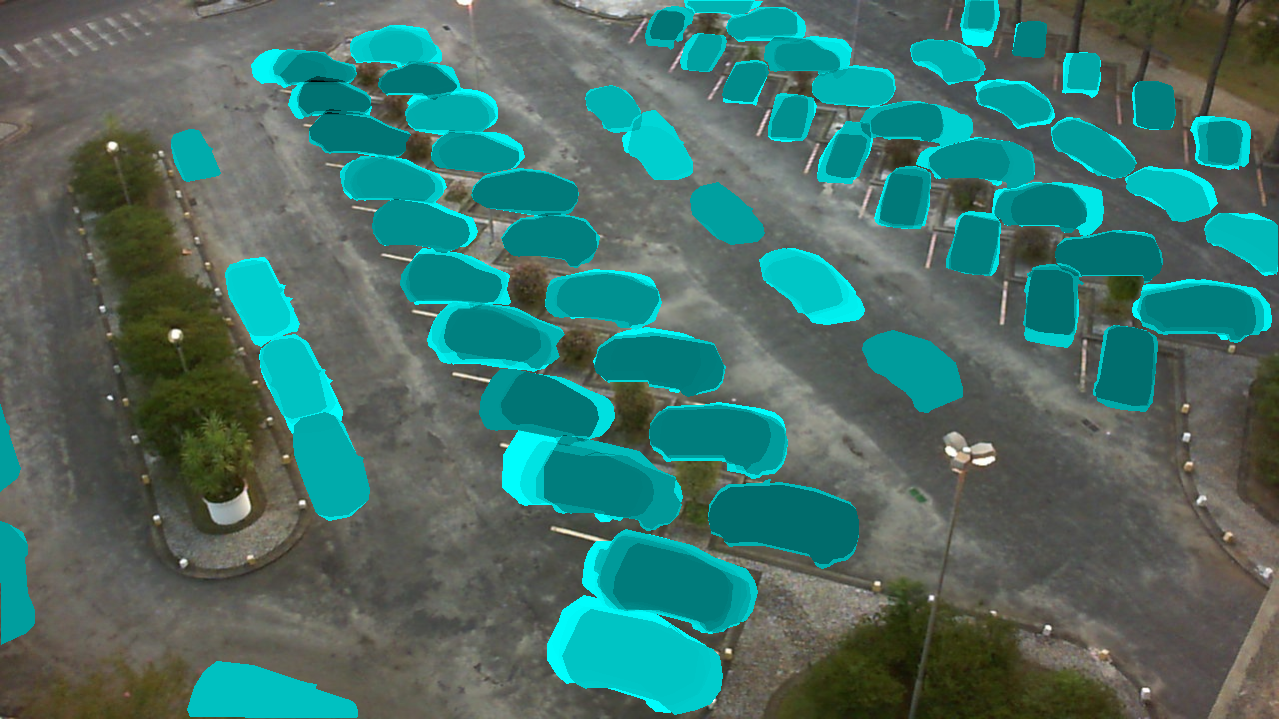}
        \caption{}                        
    \end{subfigure}            
    \begin{subfigure}[b]{0.14\textwidth}  
        \centering 
        \includegraphics[width=\textwidth]{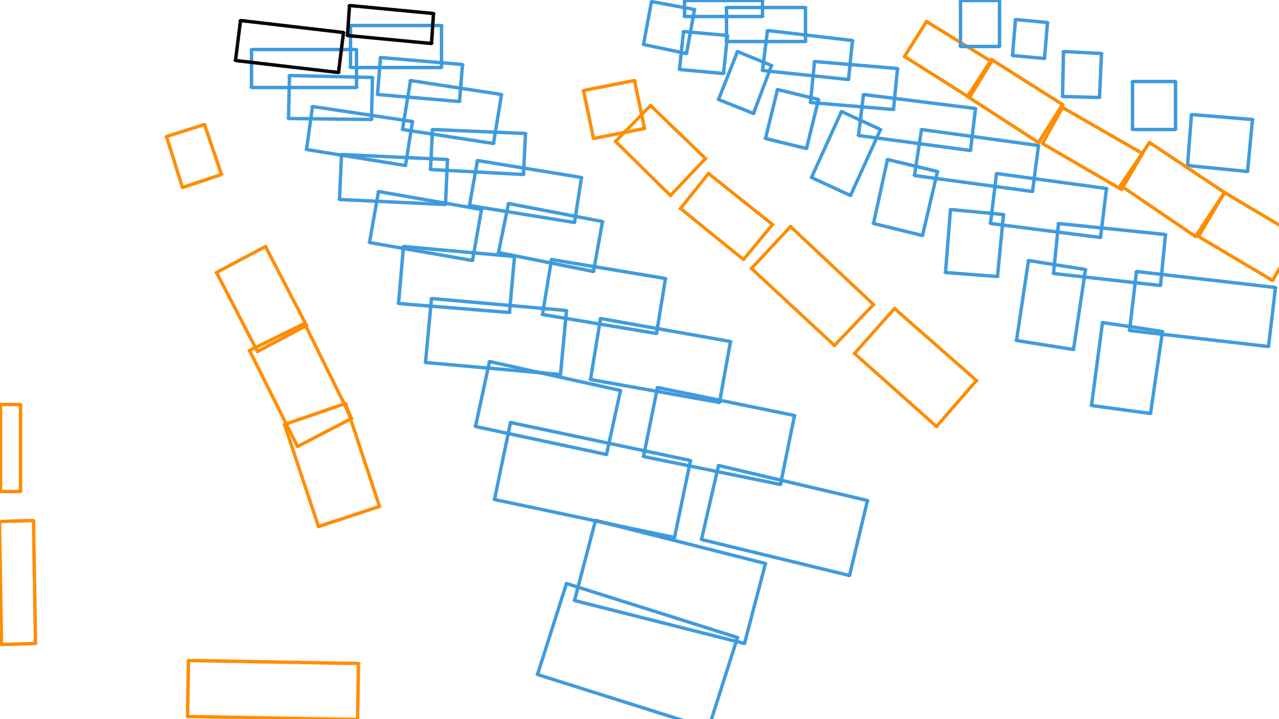}
    \caption{}                              
    \end{subfigure}
    \begin{subfigure}[b]{0.14\textwidth}  
        \centering 
        \includegraphics[width=\textwidth]{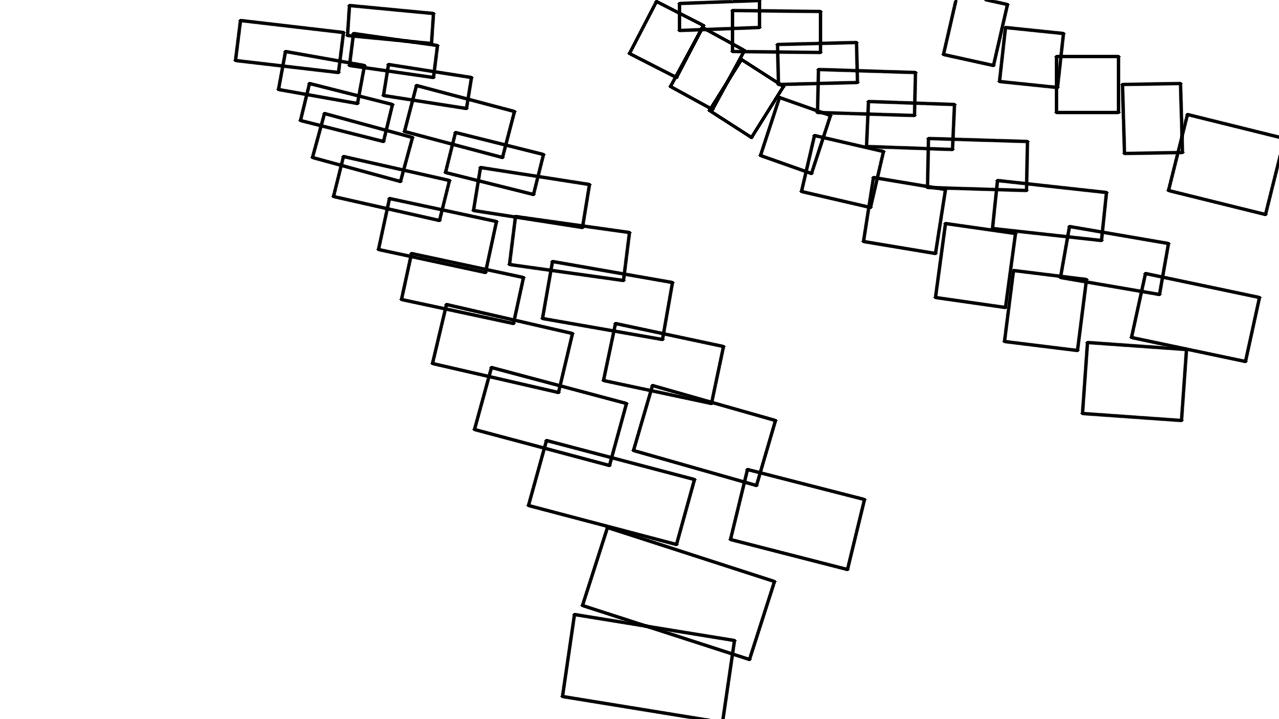}
    \caption{}                              
    \end{subfigure}
    
    \caption{Automatic Parking Space Detection from Camera3 (upper) and UFPR05 (lower part): Resulting heat map in (a,d). Automatic demarcation in (b,e), where TPs are in blue, FPs in orange, and FNs in black. Human annotation in (c,f).}
    \label{fig:result_proposed_method}
    
\end{figure}

Some examples produced by our proposed approach are shown in Figure \ref{fig:result_proposed_method}, with Camera3 on the upper part and UFPR05 on the lower part. The one-day heat map is presented in Figures \ref{fig:result_proposed_method}a and \ref{fig:result_proposed_method}d, with lighter colors indicating areas with less car occurrence and darker colors representing areas with more car occurrence. Notice that in parking spaces where vehicles are frequently moved, the pixels close to the edges are lighter, while those near the center of mass are darker due to the changing parking positions of each driver.
The resulting automatic demarcation is shown in Figures \ref{fig:result_proposed_method}b and \ref{fig:result_proposed_method}e, while the ground truth is shown in Figures \ref{fig:result_proposed_method}c and \ref{fig:result_proposed_method}f.

\begin{figure}[htpb]
\centering
    \begin{subfigure}[b]{0.18\textwidth}
        \centering
        \includegraphics[width=\textwidth]{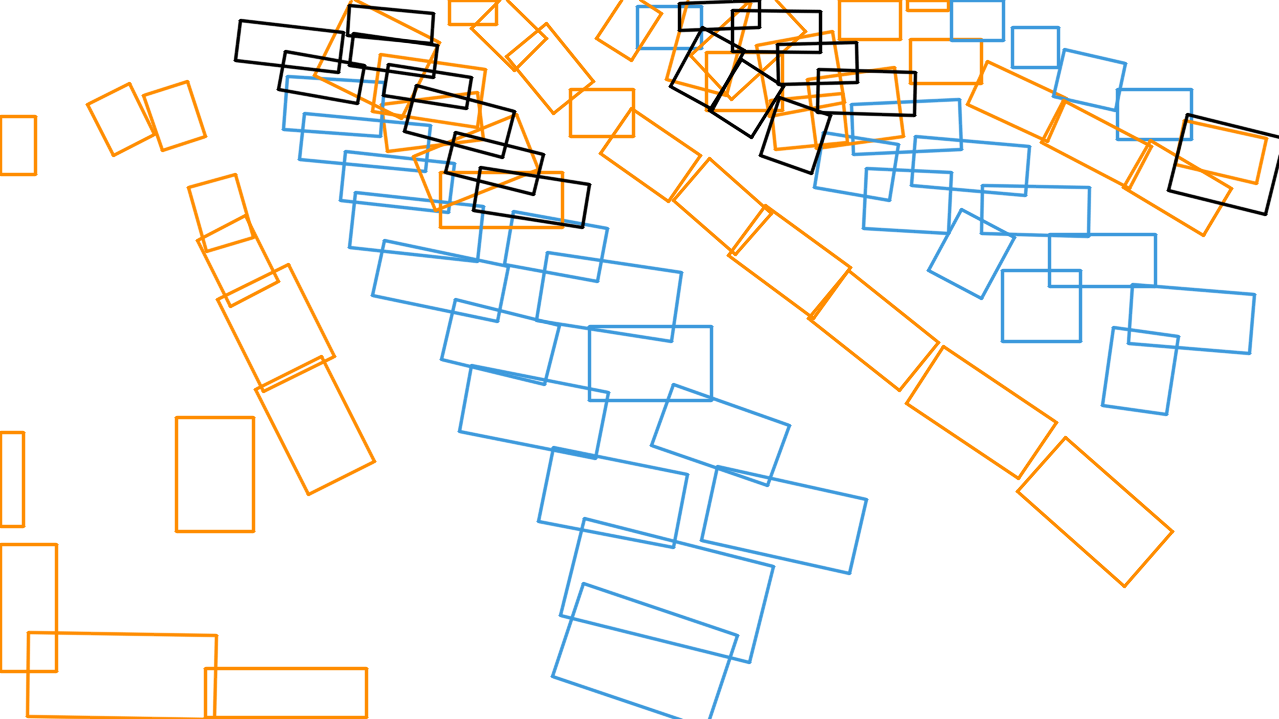}
        \caption{Results in day 2 (2013-Feb-26).}
        \label{subfig:worstResUFPR05}
    \end{subfigure}
    \hspace{0.7cm}
    \begin{subfigure}[b]{0.18\textwidth}
        \centering
        \includegraphics[width=\textwidth]{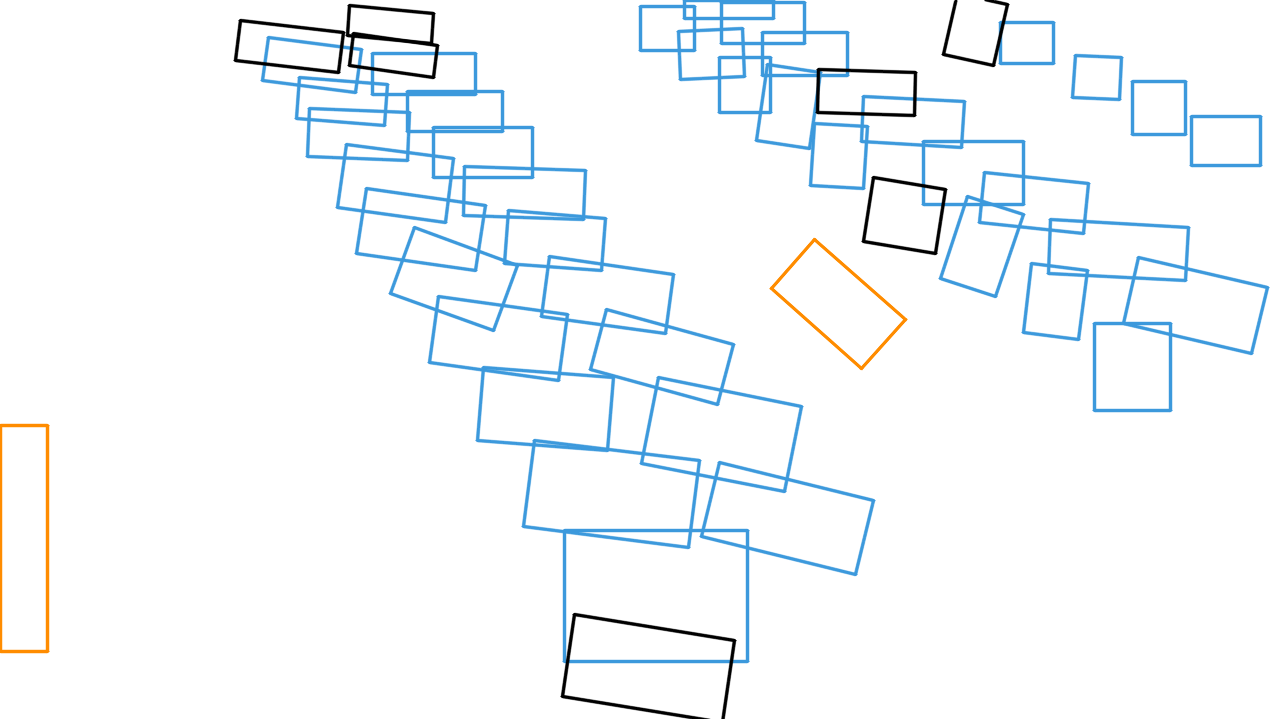}
        \caption{Results in day 17 (2013-Apr-12).}
        \label{subfig:bestResUFPR05}
    \end{subfigure}
    \vskip\baselineskip
    \begin{subfigure}[b]{0.48\textwidth}
        \centering
        \includegraphics[width=\textwidth]{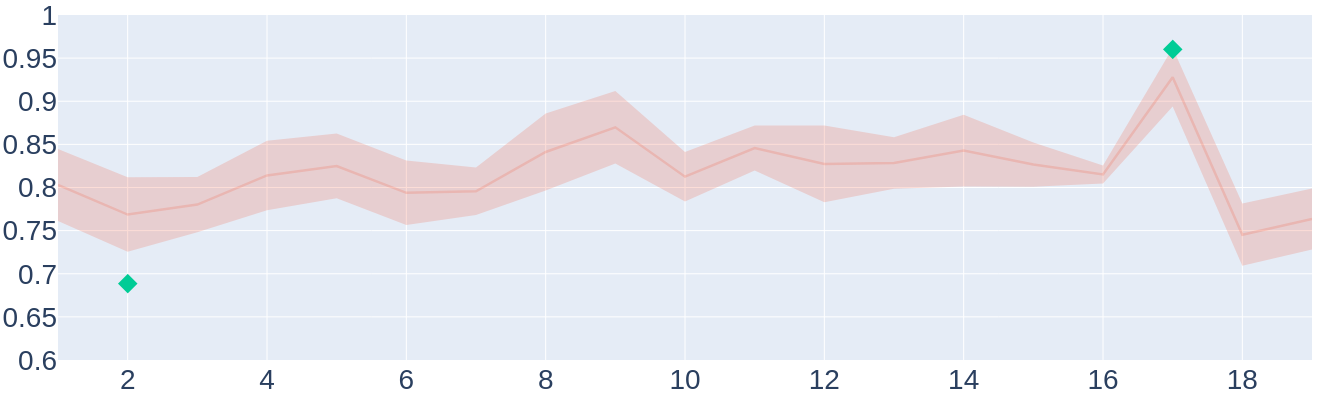}
        \caption{Mean and standard deviation of the AP25 score, in the DATASET CHANGE evaluation, for each day in UFPR05.}        
        \label{subfig:resultByDayUFPR05}
    \end{subfigure}    
    \caption{Worst (\subref{subfig:worstResUFPR05}) and (\subref{subfig:bestResUFPR05}) results achieved in the UFPR05 subset in the DATASET CHANGE evaluation. In (\subref{subfig:resultByDayUFPR05}) we show the AP25 (mean and standard deviation) achieved for every day tested. The best and worst days are highlighted in green.}
    \label{fig:illegal_parking_sensitivity}
\end{figure}

As the system relies on drivers parking cars correctly, detecting illegally parked cars will cause an over-detection of parking spaces, while unused parking spaces will not be detected. This behavior can be seen in the lower part of Figure~\ref{fig:result_proposed_method}.
Figure \ref{fig:illegal_parking_sensitivity} shows the results achieved considering each day of the UFPR05 subset. Figure \ref{fig:illegal_parking_sensitivity}c shows the average daily AP25 rates of the UFPR05 subset using the models trained on each subset from CNRPark-EXT. Figures \ref{subfig:worstResUFPR05} and \ref{subfig:bestResUFPR05} highlight the worst and best results. Most days have similar results, and the method correctly detects most legal parking spaces. The lowest performance is seen due to the number of illegally parked cars.

\begin{figure}[htpb]
    \centering
    \includegraphics[width=8cm]{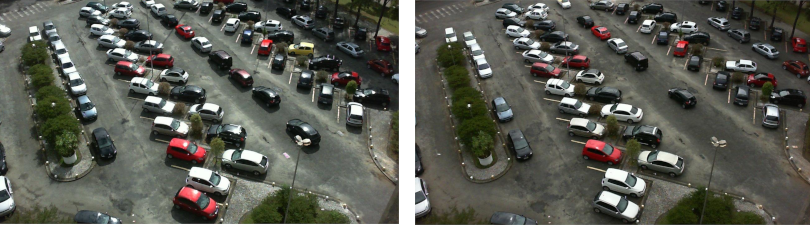}
    \caption{Samples from day 7 (2013-Mar-12) of the UFPR05 subset. The images were collected at 10:00 and 11:00 AM.}
    \label{fig:mosaic}
\end{figure}

To give a better insight into the negative impact of cars parked outside demarcated areas, we show in, Figure \ref{fig:mosaic}, sample images collected at 10:00 and 11:00 for the 7th (2013-Mar-12) day of UFPR05 subset. As one can observe, there are many cars parked irregularly. These illegally parked cars caused most of the wrong detections (see Figure \ref{fig:illegal_parking_sensitivity}).

Overall, the proposed approach reached a better AP25 score in the CNRPark-EXT than PKLot (89\% vs. 75.76\% in the ANGLE CHANGE and 85.78\% vs. 76.54\% in the DATASET CHANGE evaluations). This pattern differs from the AP50 score, where the approach reached an AP50 of 53.63\% in the PKLot, against 46.62\% of CNRPark-EXT considering the DATASET CHANGE. We hypothesize that the approach was more precise in the PKLot when detecting the real parking spaces positions. However, because of the illegally parked cars, UFPR04 and UFPR05 reduced the overall results from PKLot. Addressing this problem is crucial for better evaluating automatic parking space detectors.
\section{Conclusion}

We presented a novel approach to the problem of automatic parking space detection. Our algorithm takes as input a sequence of images of a parking lot and outputs a list of coordinates defining the detected parking spaces. The underlying assumption is that parking spaces are regions where vehicles remain parked for extended periods. The algorithm employs instance segmentation to detect parked vehicles, which is then used to create a heat map of candidate parking spaces.

We used twelve parking lot subsets from two public datasets to validate our experiments. Our experimental results show that the proposed method proved promising in detecting parking spaces in several conditions, reaching an overall AP25 of 76.54\%$\pm$12.86\% and AP50 of 53.63\%$\pm$9.62\% for the PKLot dataset and overall AP25 of 85.78\%$\pm$5.88\% and AP50 of 46.62\%$\pm$13.51\% for the CNRPark-EXT dataset, in a DATASET CHANGE scenario.

Incorrectly parked cars can significantly impact the detection results, leading to reduced AP scores in subsets such as UFPR04 and UFPR05. However, this underscores the method's adaptability to dynamic parking scenarios, including seasonal events, without requiring prior knowledge of parking demarcations. Furthermore, the method can be useful for enforcing parking fines by authorities. In future research, we will examine the impact of utilizing multiple days of operation to detect parking spaces. Additionally, we plan extend our current datasets to create more robust experimental scenarios.

\bibliographystyle{IEEEtran}
\bibliography{IEEEabrv,biblio}

\end{document}